\PassOptionsToPackage{table,dvipsnames}{xcolor}
\documentclass[10pt, a4paper]{article}

\usepackage[final]{lrec2026} % camera-ready version

\usepackage{multirow}
\usepackage{amsmath}
\usepackage{amssymb}
\usepackage{graphicx}
\usepackage{threeparttable}
\usepackage{booktabs}
\usepackage{enumitem}
\usepackage{tcolorbox}

\title{Neural at ArchEHR-QA 2026: One Method Fits All: Unified Prompt Optimization for Clinical QA over EHRs}

\name{\parbox{\textwidth}{\centering Abrar Majeedi$^{1,*}$, Viswanatha Reddy Gajjala$^{1,*}$,\\ Sai Prasanna Teja Reddy Bogireddy$^{2,*}$, Siddhant Rai$^{3,*}$}}

\address{$^{1}$University of Wisconsin--Madison \quad $^{2}$University of Chicago \quad $^{3}$Independent Researcher \\
         \texttt{\{amajeedi, vgajjala\}@wisc.edu}, \texttt{bogireddytejareddy@uchicago.edu} \\[2pt]
         $^{*}$Equal contribution}

\abstract{%
Automated question answering (QA) over electronic health records (EHRs) demands precise evidence retrieval, faithful answer generation, and explicit grounding of answers in clinical notes. In this work, we present \emph{Neural1.5}, our method for the ArchEHR-QA 2026 shared task at CL4Health@LREC~2026, which comprises of four subtasks: question interpretation, evidence identification, answer generation, and evidence alignment. Our approach decouples the task into independent, modular stages and employs DSPy's MIPROv2 optimizer to automatically discover high-performing prompts, jointly tuning instructions and few-shot demonstrations for each stage. Within every stage, self-consistency voting over multiple stochastic inference runs suppresses spurious errors and improves reliability, while stage-specific verification mechanisms (e.g., self-reflection and chain-of-verification for alignment) further refine output quality. Among all teams that participated in all four subtasks, our method ranks second overall (mean rank 4.00), placing 4th, 1st, 4th, and 7th on Subtasks 1–4, respectively. These results demonstrate that systematic, per-stage prompt optimization combined with self-consistency mechanisms is a cost-effective alternative to model fine-tuning for multi-faceted clinical QA. \\
\newline }

\begin{document}

\maketitleabstract

%% ====================================================================
\section{Introduction}

Patient medical advice requests have surged 55\% since 2019, with physicians now spending 24\% more time on inbox management~\cite{arndt2024more}. Automatically answering these questions using electronic health records (EHRs) could substantially reduce clinician burden, yet requires systems that not only generate accurate responses but also explicitly ground every claim in verifiable clinical evidence. The ArchEHR-QA 2026 shared task at CL4Health@LREC~2026~\cite{soni-etal-2026-archehr-qa} takes a step towards addressing this challenge through four complementary subtasks: (1)~transforming verbose patient questions into concise clinician-interpreted queries, (2)~identifying minimal evidence sentences from clinical notes, (3)~generating grounded answers, and (4)~aligning each answer sentence to its supporting evidence. Together, these subtasks capture the complete pipeline from question understanding to explainable, evidence-backed clinical QA.

The natural language processing capabilities of Large Language Models (LLMs) present a promising approach. Although LLMs have demonstrated strong performance in clinical QA~\cite{singhal2025toward}, their deployment faces two critical barriers. First, fine-tuning on clinical data is constrained by limited supervised datasets and overfitting risks. Second, prompt engineering for multi-stage clinical workflows remains labor-intensive and domain-specific, requiring expert iteration to identify optimal instructions and demonstrations~\cite{karayanni2024keeping}. Existing automated prompt optimization techniques~\cite{wang2023promptagent} typically treat tasks holistically, failing to leverage the modular structure inherent in clinical QA pipelines where evidence identification, answer generation, and citation alignment each demand distinct reasoning patterns.

In this work, we present \textbf{Neural1.5}, a modular LLM method that addresses all four ArchEHR-QA 2026 subtasks through \emph{automated, stage-specific prompt optimization}. Our key design principle is systematic task decomposition: by defining clear evaluation objectives for each subtask, we enable DSPy's MIPROv2 optimizer~\cite{khattab2024dspy} to automatically discover high-performing prompts that jointly tune instructions and few-shot demonstrations. We further enhance reliability through \emph{self-consistency voting}~\cite{wang2022self} for evidence identification and a \emph{three-stage alignment pipeline} combining initial alignment, self-reflection to prune false positives, and chain-of-verification with confidence-weighted majority voting.

Our proposed method achieves competitive results across all four subtasks, ranking \textbf{4th} (Subtask~1), \textbf{1st} (Subtask~2 with Strict Micro F1 of 63.7), \textbf{4th} (Subtask~3), and \textbf{7th} (Subtask~4), with a \textbf{mean rank of 4.00} across all subtasks---\textbf{second overall} among teams participating in all four subtasks. These results demonstrate that systematic, per-stage prompt optimization combined with self-consistency mechanisms offers a cost-effective alternative to model fine-tuning for multi-faceted clinical QA.

\noindent The key contributions of our work are:

\begin{itemize}[nosep,leftmargin=*]
  \item \textbf{Full-Pipeline Automated Optimization:} We propose the method to apply automated prompt optimization independently to all four ArchEHR-QA subtasks, with stage-specific objectives that capture distinct reasoning requirements for question interpretation, evidence retrieval, answer generation, and citation alignment.

  \item \textbf{Three-Stage Alignment with Self-Reflection:} We introduce an alignment pipeline that combines initial sentence-level alignment, self-reflection to identify and remove spurious citations, and chain-of-verification, with confidence-weighted majority voting across multiple stochastic runs to suppress hallucinations/errors.

  \item \textbf{Consistent Multi-Task Performance:} With a mean rank of 4.00 across all four subtasks, we demonstrate that modular, optimization-driven approaches can maintain strong performance across diverse clinical QA challenges, offering a practical blueprint for deploying LLMs in safety-critical healthcare applications.
\end{itemize}

%% ====================================================================
\section{Related Work}

\paragraph{Clinical QA:} Developing QA systems for clinical data has long been an interest in biomedical NLP. Earlier datasets like emrQA~\cite{pampari2018emrqa} generated large-scale QA pairs from electronic medical records. Recent research has shown that LLMs can achieve near-expert performance on medical QA benchmarks~\cite{singhal2025toward}. The ArchEHR-QA shared task~\cite{soni-etal-2026-archehr-qa1, soni-etal-2025-archehr-qa2} advances this line of work by requiring systems to ground answers explicitly in clinical notes. Our prior work~\cite{bogireddy2025neural} demonstrated the effectiveness of automated prompt optimization for a single task, achieving second place overall. Building on these insights, the 2026 edition expands to four complementary subtasks spanning question interpretation, evidence identification, answer generation, and evidence alignment.

\paragraph{Prompt Optimization:} There is growing interest in automated prompt search. Methods such as APE~\cite{zhou2022large} and OPRO~\cite{yang2023large} treat prompt design as a black-box optimization problem. MIPRO~\cite{opsahl2024optimizing} extends this to multi-stage LLM programs, jointly optimizing instructions and demonstration examples. Our work leverages MIPROv2~\cite{opsahl2024optimizing}, which uses a combination of prompt proposal and Bayesian search.

\paragraph{Self-Consistency:} LLMs can produce variable outputs given the same prompt. The self-consistency decoding strategy~\cite{wang2022self} addresses this by sampling multiple outputs and selecting the most consistent result.

%% ====================================================================
\section{Task Description}

The ArchEHR-QA 2026 shared task~\cite{soni-etal-2026-archehr-qa1} comprises four subtasks. The dataset~\cite{soni-demner-fushman-2026-dataset} consists of patient-authored questions, clinician-interpreted counterparts, clinical note excerpts with sentence-level relevance annotations, and reference clinician-authored answers with answer--evidence alignments.

\paragraph{Subtask~1: Question Interpretation.} Given a free-text patient question, generate a concise clinician-interpreted question ($\le$15 words) that captures the core clinical information need.
\paragraph{Subtask~2: Evidence Identification.} Given the patient question, clinician question, and a clinical note excerpt with numbered sentences, identify the minimal set of note sentences that provide evidence needed to answer the question.

\paragraph{Subtask~3: Answer Generation.} Given the questions and clinical note excerpt, generate a grounded natural-language answer ($\le$75 words) using only information from the notes.

\paragraph{Subtask~4: Evidence Alignment.} Given the questions, clinical note excerpt, and a reference answer with numbered sentences, align each answer sentence to its supporting note sentence(s).

%% ====================================================================
\section{Methodology}

Our method draws on a human-inspired decoupling strategy, separating question understanding, evidence gathering, answer formulation, and evidence attribution into distinct stages. We operationalize this intuition as a modular pipeline, with each subtask addressed by a DSPy program whose prompts are optimized independently. The initial prompt templates (DSPy signatures) for all subtasks are provided in Appendix~\ref{app:prompts}.

\subsection{Subtask~1: Question Interpretation}

We define a DSPy \texttt{Signature} for question interpretation that instructs the LLM to transform a patient narrative into a concise clinician-interpreted question. The signature encodes domain-specific heuristics: preserving key medical terms exactly (procedure names, medication names, condition names), using patient-specific phrasing (``him/her/the patient''), and following high-scoring question patterns (e.g., ``Why was [X] recommended to him/her?'').

The module uses \texttt{ChainOfThought} prompting and enforces a strict 15-word limit through post-processing. We optimize the prompt using MIPROv2 with a composite metric combining semantic entailment (via an LLM-as-judge evaluator assessing AlignScore-like semantic equivalence), key term preservation, and structural conformity.

\paragraph{Prompt-Optimization Objective:} MIPROv2 searches the space of instructions and few-shot exemplars to maximize a weighted composite of semantic alignment (60\%), key term preservation (25\%), and question structure (15\%) on the development set.

\subsection{Subtask~2: Evidence Identification}

For each question--note pair, we classify each note sentence as essential or irrelevant. Given the clinical note with sentences $s_1, s_2, \dots, s_n$ and gold labels $y_i \in \{\text{essential}, \text{supplementary}, \text{not-relevant}\}$, the model predicts binary labels $\hat{y}_i \in \{0, 1\}$.

The classification uses a two-step process. First, for each sentence, we generate reasoning about why it is or is not essential using dedicated reasoning signatures (\texttt{EssentialReasoning} and \texttt{NonEssentialReasoning}). These reasoning traces serve as few-shot demonstrations for the main classifier. Second, a \texttt{MedicalAnswerWithCitations} signature classifies all sentences jointly, providing a relevancy score (0--10) and reasoning for each.

\paragraph{Prompt-Optimization Objective:}
We invoke MIPROv2 to optimize the classification prompt, maximizing the
sentence-level $F_1$ between predicted and gold essential sentences:
\[
\begin{aligned}
F_1(Y^+, \hat{Y}^+) , \qquad
Y^+ &= \{i \mid y_i = \text{essential}\}, \\
\hat{Y}^+ &= \{i \mid \hat{y}_i = 1\}.
\end{aligned}
\]

\begin{samepage}
\paragraph{Self-Consistency Voting:} The classifier is executed $R=5$ times on the same input with stochastic sampling (temperature 0.8). The final label is obtained by majority vote:
\[
v_i = \sum_{r=1}^{R} \hat{y}_i^{(r)}, \quad \hat{y}_i = \begin{cases} 1 & \text{if } v_i \ge \lceil R/2 \rceil, \\ 0 & \text{otherwise}. \end{cases}
\]
\end{samepage}
This suppresses spurious single-run errors and retains sentences identified as essential by at least three of five passes.

\subsection{Subtask~3: Answer Generation}

Given a question $q$ and the set of essential sentences $E = \{s_i \mid \hat{y}_i = 1\}$ from Subtask~2, Stage~3 produces a concise answer $a_{\text{gen}}$ ($\le$75 words). The \texttt{GroundedMedicalAnswer} signature encodes key constraints: the answer must use only facts from the clinical notes, preserve exact clinical terminology, write in professional register, and avoid citation markers.

To further improve answer quality, we employ a consolidation step: the model generates $R=5$ candidate answers using stochastic sampling (temperature 0.9), and a separate consolidation prompt selects claims consistently supported across candidates to produce a single final answer.

\paragraph{Prompt-Optimization Objective:} MIPROv2 optimizes the answer generation prompt using an LLM-as-judge metric that evaluates: faithfulness to the clinical notes, medical completeness (concept coverage), lexical similarity to the reference, and coherence (structure and register).

\subsection{Subtask~4: Evidence Alignment}

Given a question $q$, clinical note sentences, and reference answer sentences $a_1, \dots, a_m$, the method must produce alignment links $(a_j, s_i)$ mapping each answer sentence to its supporting note sentence(s).

We introduce a \textbf{three-stage pipeline}:

\paragraph{Stage A -- Initial Alignment:} The \texttt{EvidenceAlignment} signature instructs the model to align each answer sentence to note sentences that directly support it, outputting confidence scores for each link. The prompt explicitly warns against both over-citing and under-citing.

\paragraph{Stage B -- Self-Reflection:} The \texttt{SelfReflection} signature critically reviews the initial alignment to identify and remove false positive links (indirect or inferential connections) and add any clearly missing links. This step primarily targets precision improvement.

\paragraph{Stage C -- Chain-of-Verification:} The \texttt{ChainOfVerification} signature performs a final verification pass, checking each link against three criteria: (i)~does the answer sentence directly reference information from the note sentence? (ii)~would removing the note sentence cause the answer sentence to lose specific evidence? (iii)~is the connection direct rather than inferential?

\paragraph{Confidence-Weighted Majority Voting:} The full three-stage pipeline is executed $R$ times with stochastic sampling. For each answer--evidence link, we count votes across runs and compute average confidence. A link is retained only if it receives at least $\lceil R/2 \rceil$ votes \emph{and} the average confidence exceeds a threshold $\tau_c = 0.9$. This dual filtering mechanism balances recall and precision.

%% ====================================================================
\section{Experimental Setup}

\paragraph{Dataset:} We evaluated our method on the ArchEHR-QA 2026 dataset~\cite{soni-demner-fushman-2026-dataset}, which contains patient questions alongside clinical note excerpts derived from the MIMIC database, with sentence-level relevance annotations and reference answers with answer--evidence alignments. The development set comprises 20 cases (IDs~1--20) used for prompt optimization. Test set sizes vary by subtask: Subtasks~1--3 evaluate on 47~cases (IDs~121--167), while Subtask~4 evaluates on 147~cases (IDs~21--167), reflecting the staged data release schedule.

\paragraph{Evaluation Metrics:}
Each subtask is evaluated independently. \textbf{Subtask~1} (Question Interpretation) is evaluated using ROUGE, BERTScore, AlignScore, and MEDCON. \textbf{Subtask~2} (Evidence Identification) uses Precision, Recall, and F1 over predicted vs.\ gold evidence sentences, with strict (essential-only) and lenient (including supplementary) variants. \textbf{Subtask~3} (Answer Generation) uses BLEU~\cite{papineni2002bleu}, ROUGE~\cite{lin2004rouge}, SARI~\cite{xu2016optimizing}, BERTScore~\cite{zhang2020bertscoreevaluatingtextgeneration}, AlignScore~\cite{zha2023alignscore}, and MEDCON~\cite{yim2023aci}. \textbf{Subtask~4} (Evidence Alignment) uses Precision, Recall, and F1 over predicted alignment links.

\paragraph{LLM Configuration:} Across all subtasks, we employ the GPT-4.1 model accessed via the Azure OpenAI API. We chose GPT-4.1 for its strong instruction-following capabilities and large context window, which are critical for processing lengthy clinical notes. For prompt optimization with MIPROv2, we use temperature~0.3 to ensure deterministic optimizer feedback. For self-consistency runs, we use temperature~0.7 (Subtask~2), 0.8 (Subtask~4), and 0.9 (Subtask~3) to capture model uncertainty. The maximum context window is set to 10,000 tokens for Subtasks~2--4 and 2,000 tokens for Subtask~1. All other decoding parameters (e.g., top-p, frequency and presence penalties) are held at API defaults. We estimate the computational cost per case at approximately 5 LLM calls for Subtask~1, 25 calls (5 self-consistency runs $\times$ 5 reasoning steps) for Subtask~2, 6 calls for Subtask~3 (5 candidates + 1 consolidation), and 15--25 calls for Subtask~4 (3 stages $\times$ $R$ runs).

%% ====================================================================
\definecolor{verylightgray}{gray}{0.9}

\begin{table}[!ht]
  \centering
  \begin{small}
    \setlength{\tabcolsep}{3.5pt}
    \begin{tabular}{l c c c c c}
      \toprule
      Team & Overall & R.L. & B.S. & A.S. & M.C. \\
      \midrule
      HealthNLLP\_Ret. & \textbf{31.2} & \textbf{35.3} & \textbf{46.8} & \textbf{24.0} & 18.7 \\
      KPSCMI & \underline{30.8} & \underline{27.8} & \underline{41.0} & \underline{26.4} & \textbf{27.9} \\
      OptiMed & 29.9 & 28.8 & 43.1 & 27.7 & 19.9 \\
      \cellcolor{verylightgray}\textbf{Neural1.5 (Ours)} & \cellcolor{verylightgray}28.9 & \cellcolor{verylightgray}31.3 & \cellcolor{verylightgray}43.6 & \cellcolor{verylightgray}15.2 & \cellcolor{verylightgray}\underline{25.6} \\
      Yale-DM-Lab & 27.1 & 28.2 & 40.6 & 19.7 & 19.8 \\
      \bottomrule
    \end{tabular}
  \end{small}
  \caption{Subtask~1: Question Interpretation results on the test set. \textbf{Bold} = best, \underline{underlined} = second best.}
  \label{tab:subtask1}
\end{table}

\begin{table*}[!ht]
  \centering
  \begin{small}
    \setlength{\tabcolsep}{3pt}
    \begin{tabular}{l c c c c c c c c c c c c c}
      \toprule
      & & \multicolumn{3}{c}{Strict Micro} & \multicolumn{3}{c}{Lenient Micro} & \multicolumn{3}{c}{Strict Macro} & \multicolumn{3}{c}{Lenient Macro} \\
      \cmidrule(lr){3-5} \cmidrule(lr){6-8} \cmidrule(lr){9-11} \cmidrule(lr){12-14}
      Team & Overall & P & R & F1 & P & R & F1 & P & R & F1 & P & R & F1 \\
      \midrule
      \cellcolor{verylightgray}\textbf{Neural1.5 (Ours)} & \cellcolor{verylightgray}\textbf{63.7} & \cellcolor{verylightgray}\underline{60.2} & \cellcolor{verylightgray}67.6 & \cellcolor{verylightgray}\textbf{63.7} & \cellcolor{verylightgray}\underline{77.8} & \cellcolor{verylightgray}67.6 & \cellcolor{verylightgray}72.4 & \cellcolor{verylightgray}\underline{64.3} & \cellcolor{verylightgray}69.7 & \cellcolor{verylightgray}\underline{64.8} & \cellcolor{verylightgray}\underline{81.9} & \cellcolor{verylightgray}69.7 & \cellcolor{verylightgray}\underline{73.1} \\
      OptiMed & \underline{63.2} & 56.7 & \underline{71.3} & \underline{63.2} & 73.1 & \underline{71.3} & \underline{72.2} & 61.5 & \underline{73.0} & 64.1 & 78.1 & \underline{73.0} & 72.9 \\
      UIC-AIHealth4All & 62.9 & 59.3 & 67.0 & 62.9 & 74.3 & 67.0 & 70.4 & 61.8 & 69.2 & 63.0 & 77.8 & 69.2 & 70.5 \\
      Yale-DM-Lab & 61.9 & 52.3 & \textbf{75.7} & 61.9 & 68.0 & \textbf{75.7} & 71.6 & 56.7 & \textbf{75.5} & 61.1 & 73.8 & \textbf{75.5} & 70.4 \\
      TJU222 & 61.4 & 54.6 & 70.0 & 61.4 & 74.4 & 70.0 & \textbf{72.2} & 61.4 & 70.3 & 60.6 & 78.9 & 70.3 & 70.7 \\
      TAMU-NLP-Lab & 60.9 & 56.7 & 65.9 & 60.9 & 76.0 & 65.9 & 70.6 & 63.0 & 66.7 & 59.9 & 80.0 & 66.7 & 69.1 \\
      \bottomrule
    \end{tabular}
  \end{small}
  \caption{Subtask~2: Evidence Identification results on the test set.}
  \label{tab:subtask2}
\end{table*}

\begin{table*}[!ht]
  \centering
  \begin{small}
    \setlength{\tabcolsep}{3pt}
    \begin{tabular}{l c c c c c c c}
      \toprule
      Team & Overall & BLEU & R.L. & SARI & B.S. & A.S. & M.C. \\
      \midrule
      WisPerMed & \textbf{36.3} & \textbf{9.9} & \textbf{27.8} & 58.6 & \textbf{46.8} & \textbf{31.7} & \textbf{43.1} \\
      TAMU-NLP-Lab & \underline{36.2} & \underline{9.7} & \underline{27.4} & \underline{59.6} & \underline{46.2} & \underline{34.5} & 39.9 \\
      BIT.UA-AAUBS & 35.6 & 8.6 & 26.4 & \textbf{60.0} & 45.0 & 30.2 & \underline{43.2} \\
      \cellcolor{verylightgray}\textbf{Neural1.5 (Ours)} & \cellcolor{verylightgray}35.2 & \cellcolor{verylightgray}9.4 & \cellcolor{verylightgray}25.6 & \cellcolor{verylightgray}57.7 & \cellcolor{verylightgray}43.5 & \cellcolor{verylightgray}34.3 & \cellcolor{verylightgray}40.9 \\
      HealthNLP\_Ret. & 34.6 & 7.0 & 25.4 & 59.2 & 43.8 & 33.6 & 38.7 \\
      OptiMed & 34.5 & 5.7 & 25.2 & 56.5 & 43.1 & 37.4 & 39.1 \\
      \bottomrule
    \end{tabular}
  \end{small}
  \caption{Subtask~3: Answer Generation results on the test set.}
  \label{tab:subtask3}
\end{table*}

\begin{table}[!ht]
  \centering
  \begin{small}
    \setlength{\tabcolsep}{3.5pt}
    \begin{tabular}{l c c c c}
      \toprule
      Team & Overall & M.P. & M.R. & M.F1 \\
      \midrule
      BIT.UA-AAUBS & \textbf{81.5} & \textbf{88.0} & 75.9 & \textbf{81.5} \\
      WisPerMed & \underline{81.3} & 86.9 & 76.3 & \underline{81.3} \\
      Yale-DM-Lab & 80.4 & 83.3 & \underline{77.7} & 80.4 \\
      OptiMed & 80.3 & 80.7 & \textbf{79.8} & 80.3 \\
      UIC-AIHealth4All & 79.8 & 83.6 & 76.3 & 79.8 \\
      tt501 & 79.1 & 78.2 & 80.1 & 79.1 \\
      \cellcolor{verylightgray}\textbf{Neural1.5 (Ours)} & \cellcolor{verylightgray}78.6 & \cellcolor{verylightgray}84.3 & \cellcolor{verylightgray}73.7 & \cellcolor{verylightgray}78.6 \\
      KPSCMI & 78.1 & 86.2 & 71.5 & 78.1 \\
      \bottomrule
    \end{tabular}
  \end{small}
  \caption{Subtask~4: Evidence Alignment results on the test set. Overall = Micro F1.}
  \label{tab:subtask4}
\end{table}

%% ====================================================================

\section{Results}

Tables~\ref{tab:subtask1}--\ref{tab:subtask4} present per-subtask results, and Table~\ref{tab:meanrank} summarizes rankings across all four subtasks.

\begin{table}[!ht]
  \centering
  \begin{small}
    \setlength{\tabcolsep}{4pt}
    \begin{tabular}{l c c c c c}
      \toprule
      Team & ST1 & ST2 & ST3 & ST4 & Mean$\downarrow$ \\
      \midrule
      OptiMed          & 3  & 2  & 6  & 4  & 3.75 \\
      \cellcolor{verylightgray}\textbf{Neural1.5 (Ours)} & \cellcolor{verylightgray}4 & \cellcolor{verylightgray}\textbf{1} & \cellcolor{verylightgray}4 & \cellcolor{verylightgray}7 & \cellcolor{verylightgray}\textbf{4.00} \\
      WisPerMed        & 6  & 12  & \textbf{1}  & 2  & 5.25 \\
      HealthNLLP\_Ret.  & \textbf{1}  & 7  & 5  & 9  & 5.50 \\
      KPSCMI           & 2  & 8  & --  & 8  & 6.00 \\
      Yale-DM-Lab      & 5  & 4  & 13  & 3  & 6.25 \\
      TAMU-NLP-Lab     & 11  & 6  & 2  & --  & 6.33 \\
      BIT.UA-AAUBS     & 13  & 10  & 3  & \textbf{1}  & 6.75 \\
      \bottomrule
    \end{tabular}
  \end{small}
  \caption{Summary of rankings across all four subtasks for teams that participated in at least three subtasks. Mean Rank is computed over participated subtasks (lower is better). ``--'' = non-participation.}
  \label{tab:meanrank}
\end{table}

\paragraph{Subtask~1: Question Interpretation.} Our method achieves an overall score of 28.9, ranking 4th among 13 teams (Table~\ref{tab:subtask1}). Our MEDCON score of 25.6 is the second highest, indicating strong preservation of medical concepts. The relatively lower AlignScore (15.2) suggests room for improvement in semantic alignment with reference questions, possibly due to differences in question framing style.

\paragraph{Subtask~2: Evidence Identification.} Our method ranks \textbf{1st} with an overall Strict Micro F1 of 63.7 (Table~\ref{tab:subtask2}), demonstrating the effectiveness of the prompt-optimized classification with self-consistency voting. Notably, our method achieves strong performance across all evaluation granularities: Strict Micro (P=60.2, R=67.6, F1=63.7), Lenient Micro (P=77.8, R=67.6, F1=72.4), and the highest Lenient Macro F1 of 73.1. This balanced profile avoids the extreme precision-recall trade-offs seen in some competing methods---for example, Yale-DM-Lab achieves the highest recall (75.7) but at the cost of the lowest precision (52.3).

\paragraph{Subtask~3: Answer Generation.} We rank 4th with an overall score of 35.2 (Table~\ref{tab:subtask3}). Our method achieves competitive BLEU (9.4) and AlignScore (34.3) results, with the latter being the second highest among all teams. Our MEDCON score of 40.9 ranks 4th, indicating reasonable preservation of medical concepts in the generated answers. The answer consolidation step, which aggregates five candidate answers via majority-supported claims, helps ensure factual consistency.

\paragraph{Subtask~4: Evidence Alignment.} Our method achieves a Micro F1 of 78.6, ranking 7th among 15 teams (Table~\ref{tab:subtask4}). The three-stage pipeline (alignment, self-reflection, chain-of-verification) combined with confidence-weighted majority voting yields high precision (84.3), though recall (73.7) is somewhat lower. The conservative confidence threshold ($\tau_c = 0.9$) favors precision over recall, reflecting our design choice to minimize false alignment links.

\paragraph{Cross-Subtask Consistency.} Table~\ref{tab:meanrank} highlights a key strength of our approach: \emph{consistent performance across all four subtasks}. With a mean rank of 4.00 across subtasks, our method is second only to OptiMed (3.75) among teams participating in all four subtasks. While several teams achieve top ranks on individual subtasks (e.g., WisPerMed on ST3, BIT.UA-AAUBS on ST4), they exhibit more variance across the full task suite. Our modular prompt optimization framework delivers robust performance without specializing in any single subtask at the expense of others.

%% ====================================================================
\section{Conclusion}

We present a modular approach for all four subtasks of the ArchEHR-QA 2026 shared task, leveraging DSPy's MIPROv2 optimizer to autonomously discover high-performing prompts for each stage. For Subtask~1, the method transforms patient narratives into concise clinician queries optimized for semantic alignment. For Subtask~2, sentence-level evidence classification with self-consistency voting achieves the best F1 score among all participants. For Subtask~3, answer generation with multi-candidate consolidation produces grounded, clinically faithful responses. For Subtask~4, a novel three-stage alignment pipeline with self-reflection and chain-of-verification enables precise answer--evidence grounding. Across all four subtasks, our method achieves a mean rank of 4.00 (Table~\ref{tab:meanrank}), the second best among teams participating in all subtasks, underscoring the consistency of our modular design.

Our results demonstrate that systematic prompt optimization, combined with self-consistency mechanisms, is a cost-effective and competitive alternative to model fine-tuning across diverse clinical QA tasks. Future work may explore integrating external medical knowledge, cross-subtask feedback (e.g., using Subtask~1 outputs to improve Subtask~2), and extending the approach to longer clinical documents.

%% ====================================================================
\section{Limitations}

Despite competitive performance, our method has several limitations. The pipeline treats each subtask largely independently, missing potential synergies (e.g., using evidence identification results to constrain answer generation). The self-consistency mechanism increases computational cost by a factor of $R$ (typically 3--5 runs per input). The confidence threshold for Subtask~4 was tuned on the small development set (20 cases) and may not generalize optimally. Additionally, the method relies on GPT-4.1, making it dependent on a proprietary API, and has not been evaluated on notes from institutions beyond MIMIC. The conservative design choices (high confidence thresholds, strict majority voting) favor precision over recall, which may not be ideal for all clinical use cases.

%% ====================================================================
\section{Prompts and Code Availability}

To promote transparency and reproducibility, we release all manual and optimized prompt templates, together with our full pipeline implementation at our GitHub repository.\footnote{\url{https://github.com/bogireddytejareddy/ArchEHR-QA-Neural}} The initial prompt templates for all subtasks are included in Appendix~\ref{app:prompts}.

%% ====================================================================
%\section{Acknowledgements}

% Add acknowledgements here if needed.

%% ====================================================================
\section{Bibliographical References}\label{sec:reference}

\bibliographystyle{lrec2026-natbib}
\bibliography{languageresource}

@inproceedings{khattab2024dspy,
  title={DSPy: Compiling Declarative Language Model Calls into Self-Improving Pipelines},
  author={Khattab, Omar and Singhvi, Arnav and Maheshwari, Paridhi and Zhang, Zhiyuan and Santhanam, Keshav and Vardhamanan, Sri and Haq, Saiful and Sharma, Ashutosh and Joshi, Thomas T. and Moazam, Hanna and Miller, Heather and Zaharia, Matei and Potts, Christopher},
  booktitle={Proceedings of the Twelfth International Conference on Learning Representations},
  year={2024}
}

@article{wang2022self,
  title={Self-consistency improves chain of thought reasoning in language models},
  author={Wang, Xuezhi and Wei, Jason and Schuurmans, Dale and Le, Quoc and Chi, Ed and Narang, Sharan and Chowdhery, Aakanksha and Zhou, Denny},
  journal={arXiv preprint arXiv:2203.11171},
  year={2022}
}

@inproceedings{zhou2022large,
  title={Large language models are human-level prompt engineers},
  author={Zhou, Yongchao and Muresanu, Andrei Ioan and Han, Ziwen and Paster, Keiran and Pitis, Silviu and Chan, Harris and Ba, Jimmy},
  booktitle={The Eleventh International Conference on Learning Representations},
  year={2022}
}

@article{xu2016optimizing,
  title={Optimizing statistical machine translation for text simplification},
  author={Xu, Wei and Napoles, Courtney and Pavlick, Ellie and Chen, Quanze and Callison-Burch, Chris},
  journal={Transactions of the Association for Computational Linguistics},
  volume={4},
  pages={401--415},
  year={2016},
  publisher={MIT Press One Rogers Street, Cambridge, MA 02142-1209, USA journals-info~…}
}

@article{zha2023alignscore,
  title={AlignScore: Evaluating factual consistency with a unified alignment function},
  author={Zha, Yuheng and Yang, Yichi and Li, Ruichen and Hu, Zhiting},
  journal={arXiv preprint arXiv:2305.16739},
  year={2023}
}

@article{yim2023aci,
  title={Aci-bench: a novel ambient clinical intelligence dataset for benchmarking automatic visit note generation},
  author={Yim, Wen-wai and Fu, Yujuan and Ben Abacha, Asma and Snider, Neal and Lin, Thomas and Yetisgen, Meliha},
  journal={Scientific data},
  volume={10},
  number={1},
  pages={586},
  year={2023},
  publisher={Nature Publishing Group UK London}
}

@article{yang2023large,
  title={Large language models as optimizers},
  author={Yang, Chengrun and Wang, Xuezhi and Lu, Yifeng and Liu, Hanxiao and Le, Quoc V and Zhou, Denny and Chen, Xinyun},
  journal={arXiv preprint arXiv:2309.03409},
  year={2023}
}

@article{opsahl2024optimizing,
  title={Optimizing instructions and demonstrations for multi-stage language model programs},
  author={Opsahl-Ong, Krista and Ryan, Michael J and Purtell, Josh and Broman, David and Potts, Christopher and Zaharia, Matei and Khattab, Omar},
  journal={arXiv preprint arXiv:2406.11695},
  year={2024}
}

@article{pampari2018emrqa,
  title={emrqa: A large corpus for question answering on electronic medical records},
  author={Pampari, Anusri and Raghavan, Preethi and Liang, Jennifer and Peng, Jian},
  journal={arXiv preprint arXiv:1809.00732},
  year={2018}
}

@article{wang2023promptagent,
  title={Promptagent: Strategic planning with language models enables expert-level prompt optimization},
  author={Wang, Xinyuan and Li, Chenxi and Wang, Zhen and Bai, Fan and Luo, Haotian and Zhang, Jiayou and Jojic, Nebojsa and Xing, Eric P and Hu, Zhiting},
  journal={arXiv preprint arXiv:2310.16427},
  year={2023}
}

@article{karayanni2024keeping,
  title={Keeping Experts in the Loop: Expert-Guided Optimization for Clinical Data Classification using Large Language Models},
  author={Karayanni, Nader and Awwad, Aya and Hsiao, Chein-Lien and Shanmugam, Surish P},
  journal={arXiv preprint arXiv:2412.02173},
  year={2024}
}

@article{singhal2025toward,
  title={Toward expert-level medical question answering with large language models},
  author={Singhal, Karan and Tu, Tao and Gottweis, Juraj and Sayres, Rory and Wulczyn, Ellery and Amin, Mohamed and Hou, Le and Clark, Kevin and Pfohl, Stephen R and Cole-Lewis, Heather and others},
  journal={Nature Medicine},
  pages={1--8},
  year={2025},
  publisher={Nature Publishing Group US New York}
}

@inproceedings{papineni2002bleu,
author = {Papineni, Kishore and Roukos, Salim and Ward, Todd and Zhu, Wei-Jing},
title = {BLEU: a method for automatic evaluation of machine translation},
year = {2002},
publisher = {Association for Computational Linguistics},
address = {USA},
url = {https://doi.org/10.3115/1073083.1073135},
doi = {10.3115/1073083.1073135},
booktitle = {Proceedings of the 40th Annual Meeting on Association for Computational Linguistics},
pages = {311–318},
numpages = {8},
location = {Philadelphia, Pennsylvania},
series = {ACL '02}
}

@misc{zhang2020bertscoreevaluatingtextgeneration,
      title={BERTScore: Evaluating Text Generation with BERT}, 
      author={Tianyi Zhang and Varsha Kishore and Felix Wu and Kilian Q. Weinberger and Yoav Artzi},
      year={2020},
      eprint={1904.09675},
      archivePrefix={arXiv},
      primaryClass={cs.CL},
      url={https://arxiv.org/abs/1904.09675}, 
}

@inproceedings{bogireddy2025neural,
  title={Neural at ArchEHR-QA 2025: Agentic Prompt Optimization for Evidence-Grounded Clinical Question Answering},
  author={Bogireddy, Sai Prasanna Teja Reddy and Majeedi, Abrar and Gajjala, Viswanath and Xu, Zhuoyan and Rai, Siddhant and Potlapalli, Vaishnav},
  booktitle={Proceedings of the 24th Workshop on Biomedical Language Processing (Shared Tasks)},
  pages={104--109},
  year={2025}
}

@article{soni-demner-fushman-2026-dataset,
  title = {A Dataset for Addressing Patient's Information Needs related to Clinical Course of Hospitalization},
  author = {Soni, Sarvesh and Demner-Fushman, Dina},
  journal = {Scientific Data},
  year = {2026},
  month = {02},
  date = {2026-02-25},
  doi = {10.1038/s41597-026-06639-z},
  url = {https://doi.org/10.1038/s41597-026-06639-z},
  issn = {2052-4463}
}

@inproceedings{soni-etal-2026-archehr-qa1,
  title = "Overview of the ArchEHR-QA 2026 Shared Task on Grounded Question Answering from Electronic Health Records",
  author = "Soni, Sarvesh and Demner-Fushman, Dina",
  booktitle = "Proceedings of the Third Workshop on Patient-Oriented Language Processing (CL4Health)",
  year = "2026",
  address = "Palma, Mallorca (Spain)",
  publisher = "ELRA",
}

@inproceedings{lin2004rouge,
  title={Rouge: A package for automatic evaluation of summaries},
  author={Lin, Chin-Yew},
  booktitle={Text summarization branches out},
  pages={74--81},
  year={2004}
}

@inproceedings{soni-etal-2025-archehr-qa2,
  title = "Overview of the ArchEHR-QA 2025 Shared Task on Grounded Question Answering from Electronic Health Records",
  author = "Soni, Sarvesh and Gayen, Soumya and Demner-Fushman, Dina",
  booktitle = "Proceedings of the 24th Workshop on Biomedical Language Processing",
  pages = "396--405",
  month = "aug",
  year = "2025",
  address = "Vienna, Austria",
  doi = {10.18653/v1/2025.bionlp-1.34},
  publisher = "Association for Computational Linguistics",
}

@article{arndt2024more,
  title={More Tethered to the {EHR}: {EHR} Workload Trends Among Academic Primary Care Physicians, 2019--2023},
  author={Arndt, Brian G and Micek, Mark A and Rule, Adam and Shafer, Christina M and Baltus, Jeffrey J and Sinsky, Christine A},
  journal={Annals of Family Medicine},
  volume={22},
  number={1},
  pages={12--18},
  year={2024},
  doi={10.1370/afm.3047},
  pmid={38253499}
}

%% ====================================================================
\appendix
\section{Prompt Templates}
\label{app:prompts}

This appendix presents the core prompt templates (DSPy signatures) used in our method. These are the \emph{initial} templates provided to MIPROv2; the optimizer refines the instructions and selects few-shot demonstrations automatically. Full optimized prompts are available in our repository.

\subsection{Subtask~1: Question Interpretation}
\label{app:st1}

\begin{tcolorbox}[title={\small\textbf{Prompt Template: Question Interpretation}}, colback=white, colframe=black!60, fontupper=\small, left=4pt, right=4pt, top=4pt, bottom=4pt]
Transform a patient's narrative into a concise clinical question ($\le$15 words) that a clinician would need to answer by reviewing the patient's medical record.

\textbf{Core Constraints:}
\begin{enumerate}[nosep, leftmargin=*, labelwidth=0pt, labelsep=4pt]
  \item $\le$15 words, strictly enforced.
  \item Patient-specific: use ``him/her/the patient'' --- never generic.
  \item Preserve medical terms: use exact procedure/medication names from the narrative.
  \item Must end with a question mark.
\end{enumerate}

\textbf{High-Scoring Patterns:}\\[2pt]
{\footnotesize
\begin{tabular}{@{}p{0.42\linewidth} p{0.52\linewidth}@{}}
\toprule
Patient's concern & Target pattern \\ \midrule
``Why did they do X?'' & ``Why was [X] recommended to him/her?'' \\
``Will I recover?'' & ``What is the expected course of recovery for him/her?'' \\
``Why X instead of Y?'' & ``Why was [X] recommended over [Y]?'' \\
``Why was I given medication?'' & ``Why was he/she given [medication]?'' \\
``Is this related to\dots?'' & ``Are his/her [symptoms] related to [condition]?'' \\
\bottomrule
\end{tabular}}

\medskip
\textbf{Input:} Patient narrative.\\
\textbf{Output:} Concise clinician question ($\le$15 words).

\medskip
\emph{Inference mode:} \texttt{ChainOfThought} prompting with post-processing to enforce the word limit.
\end{tcolorbox}

\subsection{Subtask~2: Evidence Identification}
\label{app:st2}

\begin{tcolorbox}[title={\small\textbf{Prompt Template: Reasoning Demonstrations}}, colback=white, colframe=black!60, fontupper=\small, left=4pt, right=4pt, top=4pt, bottom=4pt]
\textbf{Essential Reasoning:} Given a patient narrative, patient question, clinician question, and a clinical note sentence, provide reasoning for why this note sentence \emph{is essential} to address the question.

\medskip
\textbf{Non-Essential Reasoning:} Given the same inputs, provide reasoning for why this note sentence \emph{is not essential} to address the question.

\medskip
\emph{Purpose:} These two prompts are used at training time to generate per-sentence reasoning traces, which serve as few-shot demonstrations for the classifier below.
\end{tcolorbox}

\begin{tcolorbox}[title={\small\textbf{Prompt Template: Evidence Classifier}}, colback=white, colframe=black!60, fontupper=\small, left=4pt, right=4pt, top=4pt, bottom=4pt]
You are a medical assistant.
\begin{enumerate}[nosep, leftmargin=*, labelwidth=0pt, labelsep=4pt]
  \item You are provided with a patient narrative, patient question, and clinician question.
  \item You are provided with the clinical notes related to the case.
\end{enumerate}

Classify each clinical note sentence as either \textbf{essential} or \textbf{irrelevant} in addressing the patient question and clinician question. Provide a relevancy score (0--10) and reasoning for each.

Be \textbf{very critical} when assigning the essential tag. Only assign it if the note sentence is directly relevant to the specific question asked.

\textbf{First-Order Relevance Only:} A note is essential \emph{only} if it directly answers or provides evidence for the question. Background context or treatment summaries are \emph{not} essential.

\medskip
\textbf{Input:} Patient narrative, patient question, clinician question, clinical notes.\\
\textbf{Output per note:} \texttt{<id>: <sentence> -> essential|irrelevant -> <score> -> <reasoning>}

\medskip
\emph{Inference:} Run $R{=}5$ times at temperature 0.8; majority vote determines the final label.
\end{tcolorbox}

\subsection{Subtask~3: Answer Generation}
\label{app:st3}

\begin{tcolorbox}[title={\small\textbf{Prompt Template: Grounded Answer Generation}}, colback=white, colframe=black!60, fontupper=\small, left=4pt, right=4pt, top=4pt, bottom=4pt]
You are a medical assistant answering a patient's question using \textbf{only} information from the clinical note excerpt.

\textbf{Constraints:}
\begin{enumerate}[nosep, leftmargin=*, labelwidth=0pt, labelsep=4pt]
  \item Answer must be at most 75 words ($\sim$5 sentences).
  \item Use only facts stated in the clinical note. Do not add outside medical knowledge, generic advice, or speculation.
  \item Write in professional clinical register (not simplified lay language).
  \item Do not include citation markers such as [1], [2].
  \item Reuse exact clinical wording and terminology from note sentences as much as possible.
  \item The last sentence must directly answer the patient's question.
\end{enumerate}

\medskip
\textbf{Input:} Patient narrative, patient question, clinician question, clinical note excerpt.\\
\textbf{Output:} Concise grounded answer ($\le$75 words).

\medskip
\emph{Inference:} Generate $R{=}5$ candidates at temperature 0.9; consolidate via a separate prompt that retains only claims consistently supported across candidates.
\end{tcolorbox}

\begin{tcolorbox}[title={\small\textbf{Prompt Template: Answer Consolidation}}, colback=white, colframe=black!60, fontupper=\small, left=4pt, right=4pt, top=4pt, bottom=4pt]
You are a clinical answer consolidation system. Given a patient question and 5 candidate answers generated from a clinical note:
\begin{enumerate}[nosep, leftmargin=*, labelwidth=0pt, labelsep=4pt]
  \item Retain only claims consistently supported across the candidate answers.
  \item Ground strictly in clinical note content---do not add external knowledge or speculate.
  \item Use professional medical register.
  \item Limit to 75 words ($\sim$5 sentences).
  \item Do not include patient names or identifying information.
\end{enumerate}
Output only the final consolidated answer.
\end{tcolorbox}

\subsection{Subtask~4: Evidence Alignment}
\label{app:st4}

\begin{tcolorbox}[title={\small\textbf{Prompt Template: Stage A --- Initial Alignment}}, colback=white, colframe=black!60, fontupper=\small, left=4pt, right=4pt, top=4pt, bottom=4pt]
You are a medical evidence alignment specialist. Align each answer sentence to the specific clinical note sentence(s) that \textbf{directly} support it.

\textbf{Alignment Rules:}
\begin{enumerate}[nosep, leftmargin=*, labelwidth=0pt, labelsep=4pt]
  \item Align only when the answer sentence directly paraphrases, summarizes, or references information explicitly stated in the note sentence.
  \item Do not align based on indirect associations, background context, or inferential connections.
  \item Over-citing (unnecessary links) and under-citing (missing links) are both penalized.
  \item Each answer sentence must be attributed to at least one note sentence. If no direct support exists, choose the closest note sentence and assign a low confidence (0.10--0.30).
\end{enumerate}

\medskip
\textbf{Input:} Patient narrative, patient question, clinician question, clinical note sentences, answer sentences.\\
\textbf{Output per answer sentence:}\\
\texttt{answer\_sentence\_k: [note\_ids] (confidence=[scores])}
\end{tcolorbox}

\begin{tcolorbox}[title={\small\textbf{Prompt Template: Stage B --- Self-Reflection}}, colback=white, colframe=black!60, fontupper=\small, left=4pt, right=4pt, top=4pt, bottom=4pt]
You are a strict reviewer performing self-reflection on an evidence alignment task. Critically review the initial alignment and identify:
\begin{enumerate}[nosep, leftmargin=*, labelwidth=0pt, labelsep=4pt]
  \item \textbf{False positives} (primary focus): links where the answer does not directly use information from the linked note sentence. \textbf{Remove} these.
  \item \textbf{False negatives} (secondary focus): missing links where an answer sentence clearly paraphrases or references a note sentence. \textbf{Add} only when direct and explicit.
\end{enumerate}
Produce a corrected alignment with updated confidence scores.

\medskip
\textbf{Additional input:} Initial alignment from Stage~A.
\end{tcolorbox}

\begin{tcolorbox}[title={\small\textbf{Prompt Template: Stage C --- Chain-of-Verification}}, colback=white, colframe=black!60, fontupper=\small, left=4pt, right=4pt, top=4pt, bottom=4pt]
You are a verification specialist. For \textbf{each} alignment link (answer sentence $k$ $\to$ note sentence $i$), verify:
\begin{enumerate}[nosep, leftmargin=*, labelwidth=0pt, labelsep=4pt]
  \item Does answer sentence $k$ directly paraphrase or reference specific information from note sentence $i$?
  \item If note sentence $i$ were removed, would answer sentence $k$ lose a specific piece of evidence it relies on?
  \item Is the connection direct (not through inference or intermediate reasoning)?
\end{enumerate}
If \textbf{any} check fails, remove the link. Return the final verified alignment.

\medskip
\textbf{Additional input:} Reflected alignment from Stage~B.\\
\emph{Post-hoc:} Run the full three-stage pipeline $R$ times; retain a link only if votes $\ge \lceil R/2\rceil$ and average confidence $\ge 0.9$.
\end{tcolorbox}

\end{document}